\documentclass{llncs}
\usepackage{tipa}
\usepackage{arydshln}

\usepackage{graphicx}

\begin{document}

\pagestyle{empty}

\mainmatter
\author{Helen L. Bear, Richard W. Harvey, Barry-John Theobald and Yuxuan Lan}
\institute{School of Computing Sciences, University of East Anglia, Norwich, UK}
\title{Which phoneme-to-viseme maps best improve visual-only computer lip-reading?}

\titlerunning{Lecture Notes in Computer Science}

\maketitle

\begin{abstract}
A critical assumption of all current visual speech recognition systems is that there are visual speech units called visemes which can be mapped to units of acoustic speech, the phonemes.  Despite there being a number of published maps it is infrequent to see the effectiveness of these tested, particularly on visual-only lip-reading (many works use audio-visual speech).  Here we examine 120 mappings and consider if any are stable across talkers.  We show a method for devising maps based on phoneme confusions from an automated lip-reading system, and we present new mappings that show improvements for individual talkers.
\end{abstract}

\section{Introduction}
\label{sec:intro}
Phonemes are the discriminate sounds of a language~\cite{international1999handbook} and the visual equivalent, although not precisely defined, are the visemes;~\cite{chen1998audio,fisher1968confusions,Hazen1027972}. A working definition of a viseme is a set of phonemes that have identical appearance on the lips. Therefore a phoneme falls into one viseme class but a viseme may map to many phonemes: a many-to-one mapping.
In computer lip-reading there are several possibilities for phoneme-to-viseme (P2V) mappings and some are listed in, for example,~\cite{theobald2003visual} Tables 2.3 and 2.4.  Such mappings are often  consonant-only mappings \cite{binnie1976visual,fisher1968confusions,franks1972confusion,walden1977effects}; or devised from single-talker data (so are talker-dependent \cite{kricos1982differences}) or devised from highly stylised vocabularies (\cite{owens1985visemes} for example). These are useful starting points but a P2V mapping should cover all phonemes. So here we consider the possibility of using combinations of the various known mappings which cover the consonants (listed in Table~\ref{tab:consonantmappings}) and which cover vowels (Table~\ref{tab:vowelmappings}). In total we use 15 consonant maps and eight vowel maps, all of these are paired with each other to produce 120 P2V maps to test.

\section{Dataset and Data Preparation}
We use the AVLetters2 \cite{cox2008challenge} (AVL2) dataset, to train and test recognisers based upon the 120 P2V mappings. This dataset is British-English talkers reciting the alphabet seven times.  We use four talkers for training which involves tracking their faces with Active Appearance Models (AAMs)~\cite{Matthews_Baker_2004} and extracting combined shape and appearance features. We select AAM features because they are known to out-perform other feature methods in machine visual-only lip-reading~\cite{cappelletta2012phoneme}. 

\begin{figure}[!ht]
\includegraphics[width=\textwidth]{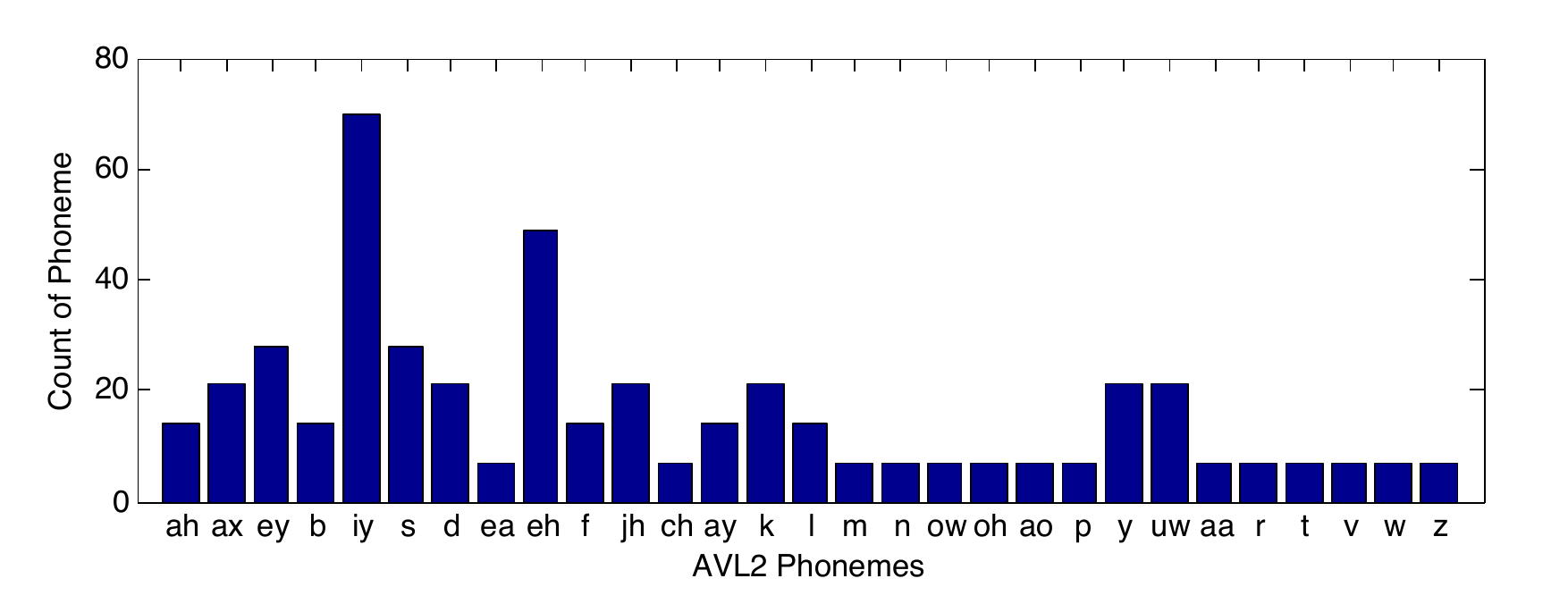}
\caption{Phoneme histogram of AVLetters-2 dataset}
\label{fig:histogram}
\end{figure}

Figure~\ref{fig:histogram} shows the count of the 29 phonemes in training component of AVL2 with the silence phoneme omitted. As is often the case, the rare phonemes in British English are not represented~\cite{cappelletta2012phoneme}. The division of these phoneme across viseme classes will vary with each different map. P2V mappings are contractive which is illustrated in Table~\ref{tab:Confusion_Factors} which lists the ratio of phonemes to visemes (excluding silence and phonemes not handled by that mapping). Thus, in Table~\ref{tab:Confusion_Factors}, the Woodward map covers 24 consonant phonemes to four viesmes and has a confusion factor (CF) of 4/24 = 0.167, whereas Jeffers vowels maps cover 23 phonemes which are mapped to eight visemes. 

\begin{table*}[t]
	\caption{Vowel Viseme:Phoneme maps}
		\begin{tabular}{ll}
		\hline
		Classification & Viseme phoneme sets \\
		\hline 
		Bozkurt \cite{bozkurt2007comparison} &  {\footnotesize \{/ei/ /\textturnv/\} \{/ei/ /e/ /\ae/\} \{/\textrevepsilon/\} \{/i/ /\textsci/ /\textschwa/ /y/\} \{/u/ /\textupsilon/ /w/\} }\\
			& {\footnotesize \{/\textscripta\textupsilon/\}  \{/\textopeno/ /\textscripta/ /\textopeno\textsci/ /\textschwa\textupsilon/\}  } \\
		Disney \cite{disney} & {\footnotesize  \{/\textupsilon/ /h/\} \{/\textepsilon\textschwa/ /i/ /ai/ /e/ /a/\} \{/u/\} \{/\textupsilon\textschwa/ /\textopeno/ /\textopeno\textschwa/\} } \\
		Hazen \cite{Hazen1027972} & {\footnotesize  \{/\textscripta\textupsilon/ /\textupsilon/ /u/ /\textschwa\textupsilon/ /\textopeno/ /w/ /\textopeno\textsci/\}  \{/\textturnv/ /\textscripta/\} \{/\ae/ /e/ /ai/ /ei/\} }\\ 
			& {\footnotesize  \{/\textschwa/ /\textsci/ /i/\} } \\
		Jeffers \cite{jeffers1971speechreading} &  {\footnotesize \{/\textscripta/ /\ae/ /\textturnv/ /ai/ /e/ /ei/ /\textsci/ /i/ /\textopeno/ /\textschwa/ /\textsci/\} \{/\textopeno\textsci/ /\textopeno/\} \{/\textscripta\textupsilon/\}  } \\ 
			& {\footnotesize \{/\textrevepsilon/ /\textschwa\textupsilon/ /\textupsilon/ /u/\} } \\
		Lee \cite{lee2002audio} & {\footnotesize   \{/i/ /\textsci/\} \{/e/ /ei/ /\ae/\} \{/\textscripta/ /\textscripta\textupsilon/ /ai/ /\textturnv/\} \{/\textopeno/ /\textopeno\textsci/ /\textschwa\textupsilon/\}   }\\
			& {\footnotesize \{/\textupsilon/ /u/\} } \\
		Montgomery \cite{montgomery1983physical} & {\footnotesize  \{/i/ /\textsci/\} \{/e/ /\ae/ /ei/ /ai/\} \{/\textscripta/ /\textopeno/ /\textturnv/\} \{/\textupsilon/ /\textrevepsilon/ /\textschwa/\}\{/\textopeno\textsci/\}  }\\
			& {\footnotesize \{/i/ /hh/\} \{/\textscripta\textupsilon/ /\textschwa\textupsilon/\} \{/u/ /u/\}  }\\
		Neti \cite{neti2000audio} & {\footnotesize  \{/u/ /\textupsilon/ /\textschwa\textupsilon/\} \{/\ae/ /e/ /ei/ /ai/\} \{/\textsci/ /i/ /\textschwa/\}  }\\
			& {\footnotesize  \{/\textopeno/ /\textturnv/ /\textscripta/ /\textrevepsilon/ /\textopeno\textsci/ /\textscripta\textupsilon/ /\textipa{H}/\} } \\
		Nichie \cite{lip_reading18} &  {\footnotesize \{/u/\} \{/\textupsilon/ /\textschwa\textupsilon/\} \{/\textscripta\textupsilon/\} \{/i/ /\textturnv/ /\textsci/\} \{/\textturnv/\} \{/i/ /\ae/\} \{/e/ /\textsci\textschwa/\}    }\\
			& {\footnotesize \{/u/\} \{/\textschwa/ /ei/\}  }\\		
		\hline
		\end{tabular}
		\label{tab:vowelmappings}
\end{table*}

\begin{table*}[!ht]
	\caption{Consonant Viseme:Phoneme maps}
		\begin{tabular}{ll}
		\hline
		Classification & Viseme phoneme sets \\
		\hline 
		Binnie \cite{binnie1976visual} &  {\footnotesize \{/p/ /b/ /m/\} \{/f/ /v/\} \{/\textipa{T}/ /\textipa{D}/\}  \{/\textipa{S}/ /\textipa{Z}/\} \{/k/ /g/\} \{/w/\}  \{/r/\} }\\
			& {\footnotesize \{/l/ /n/\} \{/t/ /d/ /s/ /z/\} }\\
			
		Bozkurt \cite{bozkurt2007comparison} &  {\footnotesize \{/g/ /\textipa{H}/ /k/ /\textipa{N}/\} \{/l/ /d/ /n/ /t/\} \{/s/ /z/\} \{/t\textipa{S}/ /\textipa{S}/ /d\textipa{Z}/ /\textipa{Z}/\}   }\\
			&  {\footnotesize \{/r/\} \{/\textipa{T}/ /\textipa{D}/\} \{/f/ /v/\} \{/p/ /b/ /m/\} }\\
		Disney \cite{disney} &  {\footnotesize \{/p/ /b/ /m/\} \{/w/\} \{/f/ /v/\} \{/\textipa{T}/\} \{/l/\} \{/d/ /t/ /z/ /s/ /r/ /n/\} }\\
			& {\footnotesize \{/\textipa{S}/ /t\textipa{S}/ /j/\} \{/y/ /g/ /k/ /\textipa{N}/\} } \\
		Finn \cite{finn1988automatic} &  {\footnotesize \{/p/ /b/ /m/\} \{/\textipa{T}/ /\textipa{D}/\} \{/w/ /s/\} \{/k/ /h/ /g/\} \{/\textipa{S}/ /\textipa{Z}/ /t\textipa{S}/ /j/\} } \\ 
		&  {\footnotesize \{/y/\} \{/z/\} \{/f/\} \{/v/\} \{/t/ /d/ /n/ /l/ /r/\} }\\
		Fisher \cite{fisher1968confusions} &  {\footnotesize \{/k/ /g/ /\textipa{N}/ /m/\} \{/p/ /b/\} \{/f/ /v/\} \{/\textipa{S}/ /\textipa{Z}/ /d\textipa{Z}/ /t\textipa{S}/\} }\\
			& {\footnotesize \{/t/ /d/ /n/ /\textipa{T}/ /\textipa{D}/ /z/ /s/ /r/ /l/\} }\\
		Franks \cite{franks1972confusion} &  {\footnotesize \{/p/ /b/ /m/\} \{/f/\} \{/r/ /w/\} \{/\textipa{S}/ /d\textipa{Z}/ /t\textipa{S}/\}  }\\
		Hazen \cite{Hazen1027972} &  {\footnotesize \{/l/\} \{/r/\} \{/y/\} \{/b/ /p/\} \{m\} \{/s/ /z/ /h/\} \{/t\textipa{S}/ /d\textipa{Z}/ /\textipa{S}/ /\textipa{Z}/\} }\\
			&   {\footnotesize \{/\textipa{N}/\} \{/f/ /v/\} \{/t/ /d/ /\textipa{T}/ /\textipa{D}/ /g/ /k/\} }\\
		Heider \cite{heider1940experimental} &  {\footnotesize \{/p/ /b/ /m/\} \{/f/ /v/\} \{/k/ /g/\} \{/\textipa{S}/ /t\textipa{S}/ /d\textipa{Z}/\} \{/n/ /t/ /d/\} } \\
			& {\footnotesize \{/l/\} \{/r/\} \{/\textipa{T}/\} }\\
		Jeffers \cite{jeffers1971speechreading} & {\footnotesize \{/f/ /v/\} \{/r/ /q/ /w/\} \{/p/ /b/ /m/\} \{/\textipa{T}/ /\textipa{D}/\} \{/t\textipa{S}/ /d\textipa{Z}/ /\textipa{S}/ /\textipa{Z}/\}  }\\
		 	&   {\footnotesize \{/g/ /k/ /\textipa{N}/\} \{/s/ /z/\} \{/d/ /l/ /n/ /t/\} }\\
		Kricos \cite{kricos1982differences} &  {\footnotesize \{/p/ /b/ /m/\} \{/f/ /v/\} \{/w/ /r/\} \{/t/ /d/ /s/ /z/\} \{/l/\}  \{/\textipa{T}/ /\textipa{D}/\}  } \\
			&  {\footnotesize \{/\textipa{S}/ /\textipa{Z}/ /t\textipa{S}/ /d\textipa{Z}/\} \{/k/ /\textipa{n}/ /j/ /h/ /\textipa{N}/ /g/\} }\\
		Lee \cite{lee2002audio} &  {\footnotesize \{/d/ /t/ /s/ /z/ /\textipa{T}/ /\textipa{D}/\} \{/g/ /k/ /n/ /\textipa{N}/ /l/ /y/ /\textipa{H}/\} \{/f/ /v/\}  }\\ 
			&  {\footnotesize \{/r/ /w/\} \{/d\textipa{Z}/ /t\textipa{S}/ /\textipa{S}/ /\textipa{Z}/\}  \{/p/ /b/ /m/\}  }\\
		Neti \cite{neti2000audio} &  {\footnotesize\{/l/ /r/ /y/\} \{/s/ /z/\} \{/t/ /d/ /n/\} \{/\textipa{S}/ /\textipa{Z}/ /d\textipa{Z}/ /t\textipa{S}/\}  \{/f/ /v/\}  }\\
			&   {\footnotesize \{/\textipa{N}/ /k/ /g/ /w/\} \{/p/ /b/ /m/\}  \{/\textipa{T}/ /\textipa{D}/\} }\\
		Nichie \cite{lip_reading18} & {\footnotesize \{/p/ /b/ /m/\} \{/f/ /v/\} \{/\textipa{W}/ /w/\}  \{/s/ /z/\} \{/\textipa{S}/ /\textipa{Z}/ /t\textipa{S}/ /j/\}  }\\
			&   {\footnotesize \{/t/ /d/ /n/\} \{/y/\} \{/\textipa{T}/\} \{/l/\}  \{/k/ /g/ /\textipa{N}/\} \{/\textipa{H}/\} \{/r/\} }\\
		Walden \cite{walden1977effects} & {\footnotesize \{/p/ /b/ /m/\} \{/f/ /v/\} \{/\textipa{T} /\textipa{D}/\} \{/\textipa{S}/ /\textipa{Z}/\} \{/w/\} \{/s/ /z/\} \{/r/\} }\\
			& {\footnotesize  \{/t/ /d/ /n/ /k/ /g/ /j/\} \{/l/\}} \\
		Woodward \cite{woodward1960phoneme} & {\footnotesize \{/t/ /d/ /n/ /l/ /\textipa{T}/ /\textipa{D}/ /s/ /z/ /t\textipa{S}/ /d\textipa{Z}/ /\textipa{S}/ /\textipa{Z}/ /j/ /k/ /g/ /h/\}}\\ 
			& {\footnotesize \{/p/ /b/ /m/\} \{/f/ /v/\} \{/w /r/ /\textipa{W}/\} } \\
		\hline
		\end{tabular}
		\label{tab:consonantmappings}
\end{table*}

We deliberately omit the following phonemes from some mappings; /si/ (Disney), /axr/ /en/ /el/ /em/ (Bozkirt), /axr/ /em/ /epi/ /tcl/ /dcl/ /en/ /gcl/ kcl/ (Hazan), and /axr/ /em/ /el/ /nx/ /en/ /dx/ /eng/ /ux/ (Jeffers) because these are American diacritics which are not appropriate for a British English phonetic dataset. Note that all 29 phonemes in AVL2 appear across the existing P2V maps, but no mapping uses all of these phonemes. Missing phonemes from a viseme map are grouped into a garbage viseme (/gar/) to ensures we measure only the performance of the previously described viseme sets. That is, we are not creating a new map by defining new visemes within an existing map.

\section{Recognition Method}
\label{sec:recognition}
Our ground truth for measuring correct recognition is a viseme transcription produced by converting a phonetic transcript of the training data to viseme labels assuming the mapping being tested (Tables~\ref{tab:consonantmappings} \&~\ref{tab:vowelmappings}). Using HTK~\cite{htk34}, we build viseme-level hidden Markov model (HMM) recognisers using with five states and five mixture components per state. We implement a leave-one-out seven-fold cross validation. Seven folds are selected as we have seven utterances of the alphabet per talker in AVL2. The HMMs are initialised using `flat start' training and re-estimated eight times and then force-aligned using HTK's \texttt{HVite}. Training is completed by re-estimating the HMMs three more times.

\section{Comparison of current P2V maps results}
\label{sec:comparison}
We measure recognition performance of the HMMs by correctness, $C$, as there are no insertion errors to consider at the word level (AVLetters 2 contains isolated words).  Correctness is measured using:
\begin{equation}
\label{eq:correctness}
C = \displaystyle \frac{N-D-S}{N}, \qquad \\
\end{equation}
where $S$ is the number of substitution errors, $D$ is the number of deletion errors and $N$ the total number of labels in the reference transcriptions.

\begin{figure*}[!htbp]
\centering
\includegraphics[width=\textwidth]{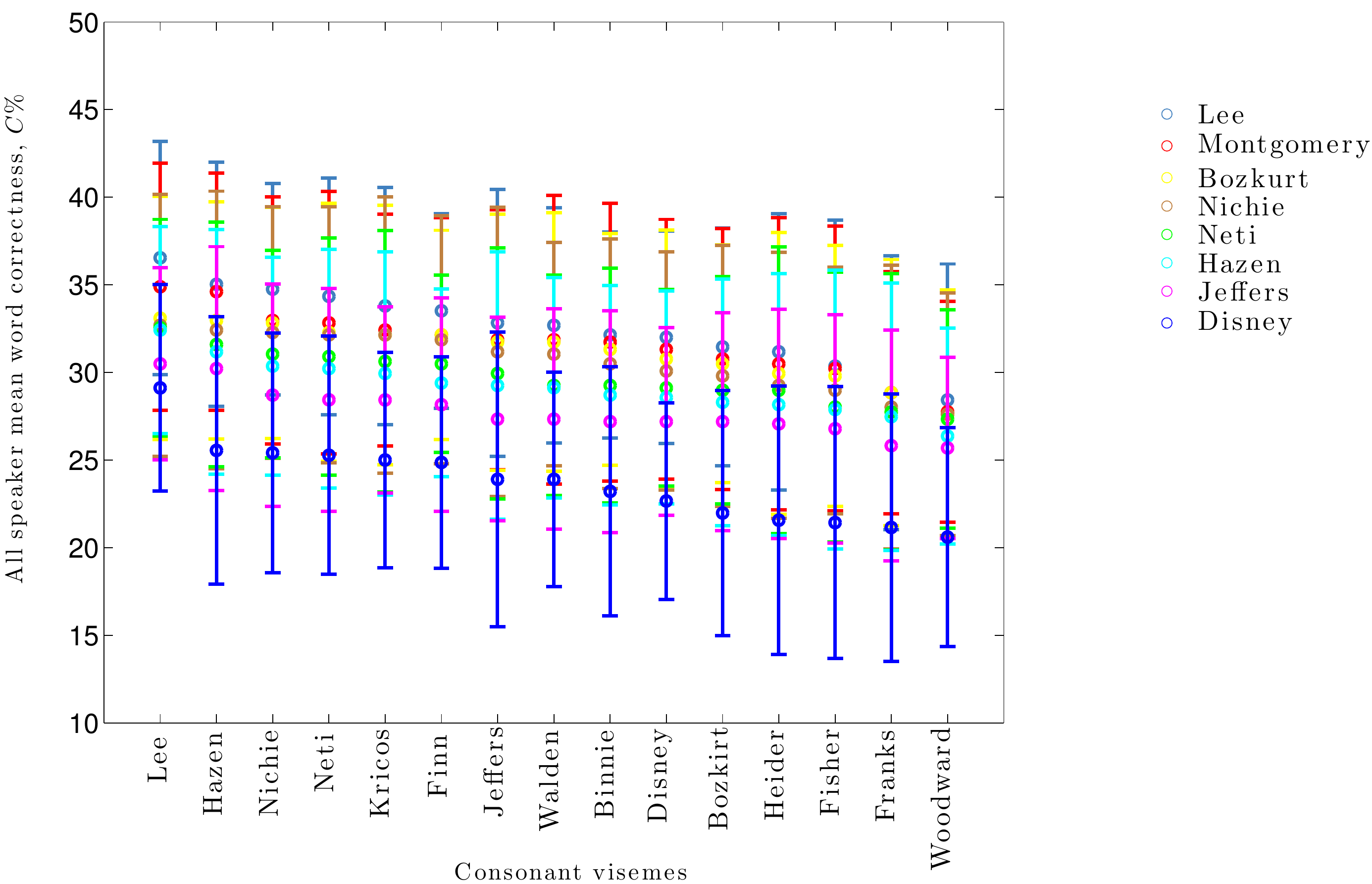} 
\caption{Talker-dependent mean word recognition $\pm$ one standard error over all four talkers comparing consonant P2V maps paired with all vowel mappings}
\label{fig:all_con}
%
%
\includegraphics[width=\textwidth]{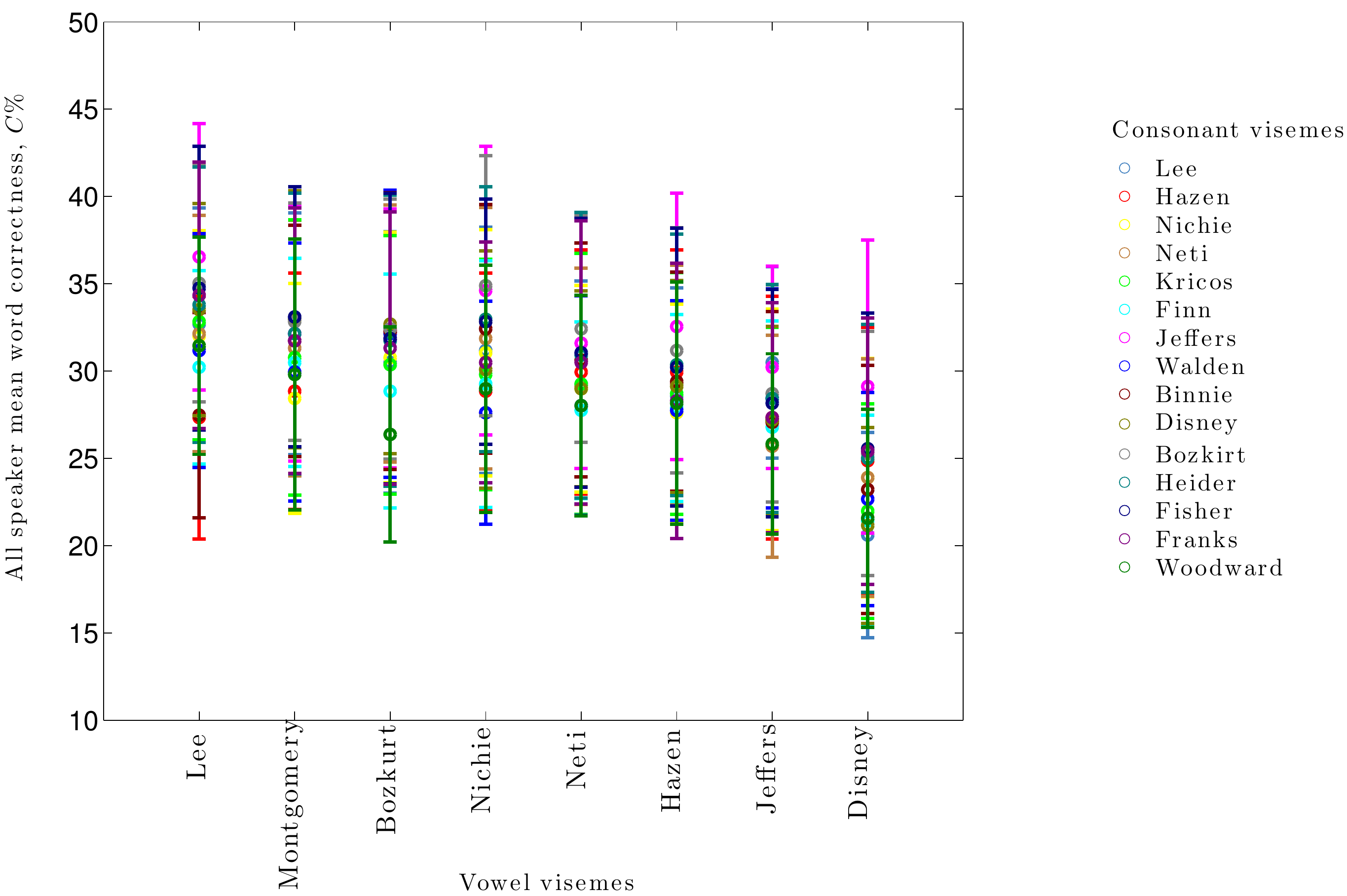} 
\caption{Talker-dependent mean word recognition $\pm$ one standard error over all four talkers comparing vowel P2V maps paired with all consonant mappings}
\label{fig:all_vow}
\end{figure*}

Word recognition is less accurate than viseme recognition. However, viseme recognition performance is not a fair test since each viseme set has a different number of visemes.  Instead, words are a common comparator that can be cross-referenced from each viseme set, and ultimately it is the difference between sets that we are interested in rather than the absolute level of performance.

Figures~\ref{fig:all_con} shows mean word correctness $\pm$ one standard error over all talkers for each consonant map along the $x$-axis paired with each vowel map.  Figure~\ref{fig:all_vow} show the same but for each vowel map along the $x$-axis paired with each consonant map. Both $x$-axes are ordered by the mean correctness. This means we can see clearly that the `best' performing map for both consonants and vowels are from Lee (as this is left-most on the $x$-axis) for all talkers.

Comparing the consonant P2V maps in Figure~\ref{fig:all_con} we see that the Disney vowels are significantly worse than all others when paired with all consonant maps. Over the other vowels there is overlap with the majority of error bars suggesting little significant difference over the whole group but Lee \cite{lee2002audio} and Bozkurt \cite{bozkurt2007comparison} vowels are consistently above the mean and above the upper error bar for Disney \cite{disney}, Jeffers \cite{jeffers1971speechreading} and Hazen \cite{Hazen1027972} vowels. In comparing the vowel P2V maps in Figure~\ref{fig:all_vow}, Lee\cite{lee2002audio} and Hazen \cite{Hazen1027972} are the best consonants by a margin above the mean whereas Woodward \cite{woodward1960phoneme} and Franks \cite{franks1972confusion} vie for bottom performance.  The best performance in terms of correctness is of a combination of vowels from Lee and consonants from Jeffers but close second best is a combination of Lee's consonants and vowels and this has a much smaller error bar.  



\begin{table}[!h]
	\caption{Confusion factors for each viseme map tested}
	\centering
	\begin{tabular}{ l  r  r  r l  r  r r}
	\hline
	Consonant Map & V:P & CF & Mean C & Vowel Map & V:P &  CF & Mean C\\
	\hline 
	Woodward  & 4:24 & 0.16 & 27.52 & Jeffers & 3:19 & 0.16 & 27.74\\
	Disney & 6:22 & 0.18 & 28.74 & Neti & 4:20 & 0.20 & 29.76\\
	Fisher & 5:21 & 0.23 & 28.86 & Hazen & 4:18 & 0.22 & 29.27\\
	Lee & 6:24 & 0.25 & 31.55 & Disney & 4:11 & 0.36 & 23.71\\
	Franks & 5:17 & 0.29 & 27.83 & Lee & 5:14 & 0.36 & 32.35\\
	Kricos & 8:24 & 0.33 & 29.46 & Bozkurt & 7:19 & 0.37 & 31.17\\
	Jeffers & 8:23 & 0.35 & 29.28 &Montgomery & 8:19 & 0.42 & 31.23\\ 
	Neti & 8:23 & 0.35 & 30.67 & Nichie & 9:15 & 0.60 & 31.13\\
	Bozkurt & 8:22 & 0.36 & 28.67 & - & - & - & -\\
	Finn & 10:23 & 0.43 & 29.43 & - & - & -& -\\
	Walden & 9:20 & 0.45 & 29.93 & - & - & -& -\\
	Binnie & 9:19 & 0.47 & 29.43 & - & - & -& -\\
	Hazen & 10:21 & 0.48 & 32.33 & - & - & -& -\\
	Heider & 8:16 & 0.50 & 28.47 & - & - & -& -\\
	Nichie & 18:33 & 0.54 & 30.94 & - & - & -& -\\

	\hline
	\end{tabular}
	\label{tab:Confusion_Factors}
\end{table} 

In Table~\ref{tab:Confusion_Factors} we present data to suggest the best performing vowel P2Vs have a ratio of phonemes-to-visemes around 0.44 (top four CF mean = 0.44), and the better performing consonant maps have a CF of approximately 0.41 (top four CF mean = 0.41) so the better P2V is $<\sim2$ phonemes per viseme.
\vspace{-0.5cm}
\section{New viseme mappings}
\label{sec:two}

Given that Lee \cite{lee2002audio} provides the best pairing of the existing phoneme to visemes maps, we now ask if there are alternatives that can perform better?
Our first approach is to find talker-dependent P2V maps based upon phoneme confusion matrices generated by a visual-only automated recognition system using phoneme HMM classifiers. Where a phoneme is only ever correctly identified as itself (true positives on the confusion matrix diagonal), this is quickly allocated to be a viseme of that single phoneme. 

Now we address the remaining phonemes which have been confused. The first candidate for viseme class 1 is a subset of Phonemes: $V_1 = \{ \phi_1, \phi_2, É \phi_{M_1} \}$ such that every pair, $(\phi_i, \phi_j)$ in $V_1$ has $N_{ij} > 0$. $V_1$ is chosen as the largest such set. $V_2$, which is the second viseme set, is determined in the same way from the remaining phonemes until all phonemes are accounted for. Within this process phonemes are grouped into a viseme class only if \emph{all} of the phonemes within the candidate group are mutually confused.  Once a phoneme has been assigned to a viseme class, it is no longer considered for grouping and so any possible other viseme combinations that include this phoneme are discarded.  

Our phoneme recognition produces confusions between consonant and vowel phonemes so we make two types of map, one that permits vowel and consonant phonemes to be mixed within the same viseme and a second which restricts visemes to be vowel or consonant phonemes only. These P2V maps for each talker are in Table~\ref{tab:tcvisemes_split}.   These are the ``tightly confused'' maps because all phonemes within each viseme have been confused with each other in the phoneme recognition.

\begin{table*}[!ht]
	\caption{Tightly confused phoneme talker-dependent visemes. The score in brackets is the ratio of phonemes to visemes}
	\begin{tabular} {ll}
	\hline 
	Classification & P2V mapping - permitting mixing of vowels and consonants\\
	\hline
	Talker1	& {\footnotesize \{/\textturnv/ /ai/ /i/ /n/ /\textschwa\textupsilon/\} \{/b/ /e/ /ei/ /y/ \} \{/d/ /s/\} \{/t\textipa{S}/ /l/\} } \\
	(CF:0.48)	& {\footnotesize  \{/t/\} \{/w/\} \{/f/\} \{/k/\} \{/\textschwa/ /v/\} \{/d\textipa{Z}/ /z/\}  \{/\textscripta/ /u/\} 	}\\
	Talker2 	& {\footnotesize\{/\textschwa/ /ai/ /ei/ /i/ /s/\} \{/e/ /v/ /w/ /y/\} \{/l/ /m/ /n/\} \{/\textturnv/ /f/\} }\\ 
	(CF: 0.44)	&  {\footnotesize \{/z/\} \/t\textipa{S}/\} \{/t/\} \{/\textscripta/\} \{/\textschwa\textupsilon/ /u/\} \{/d\textipa{Z}/ /k/\} \{/b/ /d/ /p/\}  }	\\
	Talker3 	& {\footnotesize \{/ei/ /f/ /n/\} \{/d/ /t/ /p/\} \{/b/ /s/\} \{/l/ /m/\} \{/\textschwa/ /e/\} \{/i/\} }\\
	(CF: 0.68)	& {\footnotesize \{/\textscripta/\} \{/d\textipa{Z}/\} \{/\textschwa\textupsilon/\} \{/z/\} \{/y/\} \{/t\textipa{S}\}/ \{/ai/\} \{/\textturnv/\} \{/\textscripta/\} \{/d\textipa{Z}/\} } \\
			& {\footnotesize \{/k/ /w/\}  \{/\textschwa\textupsilon/\} \{/z/\} \{/v/\} \{/u/\} 	 } \\
	Talker4	& {\footnotesize \{/\textturnv/ /ai/ /i/ /ei/ \} \{/m/ /n/\} \{/\textschwa/ /e/ /p/\} \{/k/ /w/\} \{/d/ /s/\} } \\ 
	(CF: 0.64)	& {\footnotesize \{/f/\} \{/v/\} \{/\textscripta/\} \{/z/\} \{/t\textipa{S}/\} \{/b/\} \{/\textschwa\textupsilon/\}	\{/d\textipa{Z}/ /t/\} \{/b/\} \{/\textschwa\textupsilon/\}	}\\
			& {\footnotesize  \{/l/\} \{/u/\} } \\
	\hline
	Classification & P2V mapping - restricting mixing of vowels and consonants \\
	\hline 
	Talker1	& {\footnotesize \{/\textturnv/ /i/ /\textschwa\textupsilon/ /u/\} \{/\textscripta/ /ei/\} \{/\textschwa/ /e/ /ei/\} \{/d/ /s/ /t/ \} \{/t\textipa{S}/ /l/\} }\\
	(CF:0.50)	& {\footnotesize \{/k/\} \{/z/\} \{/w/\} \{/f/\}  \{/m/ /n/\} \{/d\textipa{Z}/ /v/\} \{/b/ /y/\} 	}\\
	Talker2 	& {\footnotesize \{/ai/ /ei/ /i/ /u/\} \{/\textschwa\textupsilon/\} \{/\textschwa/\} \{/e/\} \{/\textturnv/\} \{/\textscripta/\} \{/v/ /w/\} \{/k/\}  }\\ 
	(CF: 0.58)	& {\footnotesize  \{/d/ /b/\} \{/t/\} \{/t\textipa{S}/\} \{/l/ /m/ /n/\} \{/d\textipa{Z}/ /p/ /y/\} \{/f/ /s/\}  }\\
	Talker3 	& {\footnotesize \{/ei/ /i/\} \{/ai/\} \{/\textschwa/ /e/\} \{/\textturnv/\} \{/d/ /p/ /t/\} \{/l/ /m/\} \{/k/ /w/\}  }\\
	(CF: 0.68)	& {\footnotesize \{/t\textipa{S}/\} \{/\textschwa\textupsilon/\} \{/y/\} \{/u/\} \{/\textscripta/\} \{/z/\}  \{/b/ /s/\} \{/v/\} \{/d\textipa{Z}/\} }\\
			& {\footnotesize \{/f/ /n/\} } \\
	Talker4 	& {\footnotesize \{/\textturnv/ /ai/ /i/ /ei/\} \{/\textschwa/ /e/\} \{/m/ /n/\} \{/k/ /l/\} \{/d\textipa{Z}/ /t/\}  \{/b/\} }\\ 
	(CF: 0.65)	& {\footnotesize  \{/\textschwa\textupsilon/\} \{/y/\} \{/u/\} \{/\textscripta/\} \{/w/\} \{/f/\} \{/v/\} \{/t\textipa{S}/\} \{/d/ /s/\} }\\
	\hline
	\end{tabular}
	\label{tab:tcvisemes_split}
\end{table*}


These viseme sets will contain spurious phonemes that cannot be grouped into a viseme because they are not confused with \emph{all} of the phonemes of the viseme. This leaves some single-phoneme visemes (e.g. /u/ in Talker 1 with mixed vowel and consonant phonemes), so our second approach relaxes the condition requiring confusion with all of the phonemes. We execute a second pass through the viseme sets.  Any single-phoneme viseme classes are then permitted to merge with existing multi-phoneme classes if they share any confusions with that class.  In the event that a phone has multiple class confusions it is merged with the class with the greatest confusion. We term these the ``loosely confused'' maps. Again we do two sets with vowel and consonant phonemes both mixed and separate. The final P2V maps are in Table~\ref{tab:lcvisemes_split} for four talkers.

\begin{table*}[!ht]
	\caption{Loosely confused phoneme talker-dependent visemes. The score in brackets is the ratio of phonemes to visemes}
	\begin{tabular} {ll}
	\hline 
	Classification & P2V mapping - permitting mixing of vowels and consonants \\
	\hline
	Talker1 	& {\footnotesize \{/b/ /e/ /ei/ /p/ /w/ /y/ /k/\} \{/\textturnv/ /ai/ /f/ /i/ /m/ /n/ /\textschwa\textupsilon/\} }\\
	(CF:0.28)	& {\footnotesize \{/d\textipa{Z}/ /z/\} \{/\textscripta/ /u/\}	  \{/d/ /s/ /t/\} \{/t\textipa{S}/ /l/\}  \{/\textschwa/ /v/\}\{/\textschwa/ /v/\} }\\
	Talker2 	& {\footnotesize \{/\textscripta/ /\textschwa/ /ai/ /ei/ /i/ /s/ /t\textipa{S}/\} \{/e/ /t/ /v/ /w/ /y/\} \{/l/ /m/ /n/\} }\\ 
	(CF: 0.32)	& {\footnotesize  \{/\textturnv/ /f/\} \{/z/\}  \{/b/ /d/ /p/\} \{/\textschwa\textupsilon/ /u/\} \{/d\textipa{Z}/ /k/\}	}\\
	Talker3	& {\footnotesize \{/\textturnv/ /ai/ /ei/ /f/ /i/ /n/\} \{/\textschwa/ /e/ /y/ /t\textipa{S}/\} \{/b/ /s/ /v/\} 	}\\
	(CF: 0.40)	& {\footnotesize  \{/d\textipa{Z}/\} \{/\textschwa\textupsilon/\} \{/z/\} \{/l/ /m/ /u/\} \{/d/ /p/ /t/\} \{/k/ /w/\} \{/\textscripta/\} }\\
	Talker4 	& {\footnotesize \{/\textturnv/ /ai/ /t\textipa{S}/ /i/ /ei/ \} \{/\textscripta/ /m/ /u/ /n/\} \{/\textschwa/ /e/ /p/ /v/ /y/\}  }\\ 
	(CF: 0.32)	& {\footnotesize \{/d\textipa{Z}/ /t/\} \{/k/ /l/ /w/\} \{/\textschwa\textupsilon/\} \{/d/ /f/ /s/\} \{/b/\} 	}\\
	\hline
	Classification & P2V mapping - restricting mixing of vowels and consonants\\
	\hline 
	Talker1 	& {\footnotesize \{/\textturnv/ /i/ /\textschwa\textupsilon/ /u/\} \{/\textscripta/ /ai/\} \{/\textschwa/ /e/ /ei/\} \{/b/ /w/ /y/\} }\\
	(CF:0.47)	& {\footnotesize \{/k/\} \{/z/\} \{/m/\} \{/l/\} \{/d/ /f/ /s/ /t/\} \{/t\textipa{S}/\}   \{/d\textipa{Z}/ /k/ /v/ /z/\} 	}\\
	Talker2 	& {\footnotesize \{/\textscripta/ /\textturnv/ /\textschwa/ /ai/ /ei/ /i/ /\textschwa\textupsilon/ /u/\} \{/k/ /t/ /v/ /w/\}  }\\ 
	(CF: 0.29)	& {\footnotesize \{/f/ /s/\} \{/t\textipa{S}/ /l/ /m/ /n/\} \{/d\textipa{Z}/ /p/ /y/\}  \{/b/ /d/\} \{/z/\}	}\\
	Talker3 	& {\footnotesize \{/\textturnv/ /ai/ /i/ /ei/\} \{/\textschwa/ /e/\} \{/b/ /s/ /v/\} \{/d/ /p/ /t/\}}\\ 
	(CF: 0.56)	& {\footnotesize \{/y/\} \{/d\textipa{Z}/\} \{/\textschwa\textupsilon/\} \{/z/\} \{/u/\} \{/\textschwa/ /e/\}  \{/l/ /m/\} \{/k/ /w/\} }	\\	
			& {\footnotesize  \{/f/ /n/\}  \{/\textscripta/\}   \{/t\textipa{S}/\} } \\
	Talker4 	& {\footnotesize \{/\textturnv/ /ai/ /i/ /ei/\}  \{/t\textipa{S}/ /k/ /l/ /w/\} \{/d/ /f/ /s/ /v/\} \{/m/ /n/\}  }\\ 
	(CF: 0.50)	& {\footnotesize \{/f/\} \{/\textscripta/\} \{/d\textipa{Z}/ /t/\} \{/\textschwa\textupsilon/\} \{/u/\} \{/y/\} \{/b/\} 	}\\
	\hline
	\end{tabular}
	\label{tab:lcvisemes_split}
\end{table*}


Looking at Tables~\ref{tab:tcvisemes_split} and~\ref{tab:lcvisemes_split} there are no identical visemes with each map type between talkers, this confirms our variability of individual talker visual speech (excluding the true positive single phoneme visemes). We observe that none of the new visemes match the previously suggested visemes in the comparison study (Tables~\ref{tab:vowelmappings} and~\ref{tab:consonantmappings} e.g. the most common previous viseme was \{/p/ /b/ /m/\} and this is never created with our new method.

\begin{figure}[!ht]	
\centering
	\includegraphics[width=\textwidth]{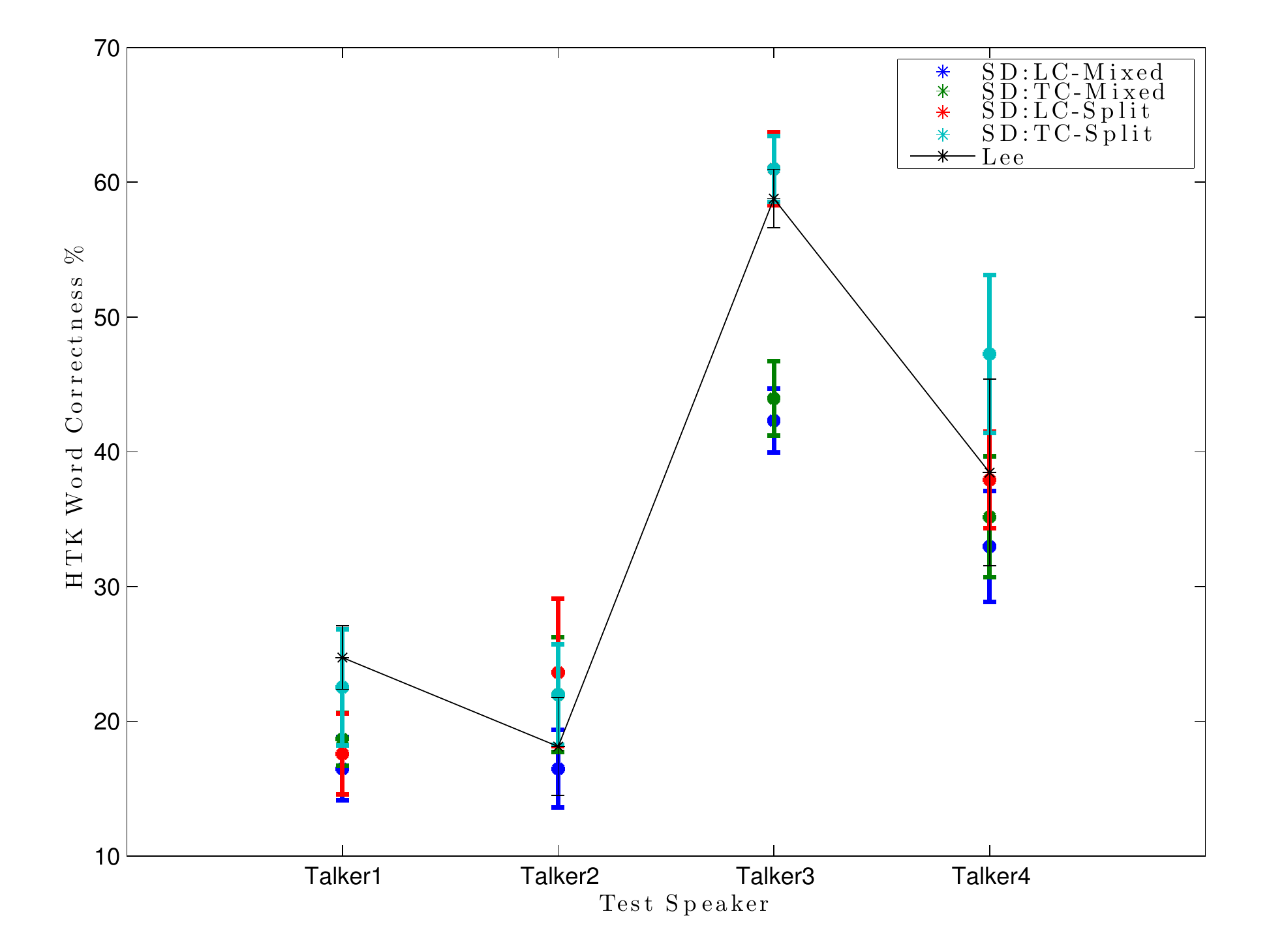} 
	\caption{HTK word Correctness using tightly confused and loosely confused viseme sets based on phoneme recognition confusions. SD = Speaker Dependent, LC = Loosely coupled, TC = Tightly Coupled, Mixed = Mixed vowels and consonant phonemes within viseme classes and Split = separated vowel and consonant visemes}
	\label{fig:q1results}
\end{figure}

Figure~\ref{fig:q1results} shows the word recognition performance using both the tightly confused map and the loosely confused map for each talker.  Also shown is the performance using the Lee map as a benchmark.  For Talker 1 no new viseme map significantly improves upon the benchmark performance, but we do see significant improvements for both Talker 2 and Talker 4 and a minor improvement within the error bars for Talker 3.  For Talkers 2 and 3, both types of the split vowels and consonant maps demonstrate improvement on the benchmark, and for Talker 4 the tightly confused split vowels and consonants shows a significant improvement.  Comparing mixed consonant and vowel maps against split consonant and vowel maps, the split maps are always better than mixed maps for all talkers in this data. In comparing the loosely confused maps versus the tightly confused maps, the tight confusions are better for two out of our four talkers (Talkers 1 and 2) and equal for a third (Talker1). These are talkers with highest confusion factor P2V maps (Tables~\ref{tab:tcvisemes_split} \&~\ref{tab:lcvisemes_split}). This is despite the tightly confused viseme set including single phoneme-viseme classes which can be confused with parts of the tightly confused classes.

\section{Conclusions and Future work}

We have completed a comprehensive experimental study of previously suggested P2V maps and shown that Lee~\cite{lee2002audio} is the best of the previously published P2V maps.  Puzzlingly the Lee mapping is not that popular among engineers of lip-reading systems so our finding should be of immediate use.

We have also outlined how it is possible to build phoneme-to-viseme maps in a systematic way using confusion matrices from real recognisers.  We believe that this is a more principled approach than previous methods (including Lee's~\cite{lee2002audio} whose method is bound by the Fisher~\cite{fisher1968confusions} visemes) and also allows comparison between talkers using phonetic terminology.  Further we have shown that the automatic method need do no worse than the Lee visemes and can exceed performance.  We acknowledge that our dataset is still rather small and the sparsely represented phonemes are unlikely to be accurately modelled.  In future we would like to extend this to full set of American and British phonemes but that will require a more extensive set of data.

\bibliographystyle{splncs}
\bibliography{icip2014}

\end{document}